# The Sparse Reverse of Principal Component Analysis for Fast Low-Rank Matrix Completion

Abdallah Chehade, *Member, IEEE* and Zunya Shi

*Abstract*—Matrix completion constantly receives tremendous attention from many research fields. It is commonly applied for recommender systems such as movie ratings, computer vision such as image reconstruction or completion, multi-task learning such as collaboratively modeling time-series trends of multiple sensors, and many other applications. Matrix completion techniques are usually computationally exhaustive and/or fail to capture the heterogeneity in the data. For example, images usually contain a heterogeneous set of objects, and thus it is a challenging task to reconstruct images with high levels of missing data. In this paper, we propose the sparse reverse of principal component analysis for matrix completion. The proposed approach maintains smoothness across the matrix, produces accurate estimates of the missing data, converges iteratively, and it is computationally tractable with a controllable upper bound on the number of iterations until convergence. The accuracy of the proposed technique is validated on natural images, movie ratings, and multisensor data. It is also compared with common benchmark methods used for matrix completion.

## I. Introduction

**M**ATRIX COMPLETION is a common task for estimating missing data in matrices [1]–[5]. It constantly receives tremendous attention from many research fields such as collaborative filtering (e.g., recommender systems) [6], [7], link analysis [8], distance embedding [3], computer vision [9], image processing [10], [11], and so forth.

During the last decade, the rapid development of sensing and information technologies has provided rich and heterogeneous data environments. This provided unprecedented opportunities for (i) accurate reliability analysis [12]–[15], (ii) accurate recommendations (advertisements, movies, stocks) [6], [7], [16], (iii) automated decision-making [17], and (iv) transfer learning between data from different task domains (spectrum, images, text documents) [18], [19]. However, in the era of Big Data and the Internet of Things, there are high volumes of missing data, which may negatively impact various data analysis tasks [20]. Therefore, it is necessary to develop effective methods that recover the missing data of interest. In comparison with traditional statistical models designed for missing value estimation, matrix completion often performs better because it accounts for similarities between the matrix entities with no strong assumptions on the distribution of the entities [20]. Consequently, scalable and novel algorithms for matrix completion are still in constant demand, especially for applications with high levels of missing data.

Generally, there are two schemes of matrix completion, with different assumptions about the rank of the matrix being recovered (low-rank and high-rank matrix completion). At present, there is more literature on low-rank matrix completion than on high-rank matrix completion. This is mainly because most data matrices analyzed are low-rank or approximately low-rank structured [3]. Taking a movie recommender system as an example, there are only a few factors that may contribute to users' preferences. This suggests that the data matrix recording users' rating scores is actually low-rank structured.

Approaches that solve low-rank matrix completion problems can be mainly divided into two categories: nuclear norm based and matrix factorization based [4]. In the first category, the objective of rank minimization is approximated by nuclear norm minimization. Then, it is able to reconstruct the original matrix with low-rank through methods like interior-point-based semidefinite programming (SDP) solver [3], [21], conjugate gradient method [22], singular value threshold (SVT) algorithm [1], augmented Lagrange multiplier (ALM) algorithm [23], other robust principal component analysis [24], [25], etc. In the second category, the original matrix is compactly represented as the product of two low-rank matrices. The two low-rank matrices are usually iteratively updated through various algorithms such as alternating least squares (ALS) [26], [27] and stochastic gradient descent (SGD) [28]–[31].

Furthermore, the most common method for high-rank matrix completion is to first separate the data matrix into a number of subspaces and then complete the multi-subspace matrices via low-rank approaches [1], [32].

Even though there is a long list of techniques availablefor matrix completion, most of them heavily rely on the concept of preserving the original observed part of the matrix. Yet there is a very important requirement for a realistically and practically good matrix completion, which is the local and global smoothness in the reconstructed matrix. For example, in image processing, it is critical to obtain smoothness over the image. Often, this requirement comes at the cost of moderately updating the observed values (e.g., Gaussian filters and Variational Bayesian techniques).

To address this limitation, we propose the Sparse Reverse of the PCA (SRPCA) to efficiently complete matrices. The

A. Chehade is with the Department of Industrial and Manufacturing Systems Engineering, University of Michigan-Dearborn, Dearborn, MI 48128 USA (e-mail: achehade@umich.edu).

Z. Shi is with the Department of Industrial and Manufacturing Systems Engineering, University of Michigan-Dearborn, Dearborn, MI 48128 USA.

proposed approach (i) maintains a high level of smoothness by iteratively finding the principal components of the matrix based on the predicted values of both the missing and the observed parts of the matrix, while (ii) guaranteeing that the principal components are capable of reconstructing the observed part of the matrix with minimal differences. Another major contribution of the paper is proposing a new perspective on utilizing PCA for sparse matrices, which is a common tool.

The remainder of the paper is organized as follows. Section II provides a general literature review of matrix completion and principal component analysis. Section III proposes the SRPCA approach for the matrix completion problem under extreme sparse conditions. Section IV evaluates the performance of the proposed approach in comparison to existing benchmark approaches in three areas (images, movie ratings, and multisensor data). Section V provides a conclusion and a discussion of potential future research directions.

## II. LITERATURE REVIEW

This section reviews common approaches used for the low-rank matrix completion problem and also reviews the concept behind PCA.

### A. Matrix Completion

If we assume that the data matrix to be recovered is a low-rank structure, the matrix completion problem should be defined as follows [3]:

$$\min_{M} \text{rank}(M) \quad \text{s.t.} \; x_{ij} = m_{ij}, \forall (i,j) \in \Omega, \quad (1)$$

where $X \in \mathbb{R}^{m \times n}$ is the sparse observed matrix, $M \in \mathbb{R}^{m \times n}$ is the reconstructed matrix of $X$, and $\Omega$ represents the observed entries of $X$.

This problem is a simple explanation of the low-rank matrix completion problem. Unfortunately, the rank minimization is NP-hard and has led researchers to propose different relaxations to solve the problem. Specifically, a commonly used convex relaxation for the rank is the nuclear norm, $\|M\|_*$, which approximates problem (1) as [3], [4], [27]:

$$\min_{M} \|M\|_*, \quad \text{s.t.} \; x_{ij} = m_{ij}, \forall (i,j) \in \Omega \quad (2)$$

or,

$$\min_{M} \tau\|M\|_* + \frac{1}{2}\|X - M\|_F^2. \quad (3)$$

Problems (2) and (3) can be conveniently optimized through some interior-point-method-based SDP solvers [3] like SDPT3 and SeDuMi. In addition, Jian-Feng Cai et al. [25] further proposed an SVT algorithm to solve problems (2) and (3). In the SVT, the estimate $M$ converges to a unique solution of problem (3) through an iterative algorithm designed as follows:

$$\begin{cases} M^{(k)} = D_\tau(M_1^{(k-1)}), \\ M_1^{(k)} = M_1^{(k-1)} + \delta_k \mathcal{P}_\Omega(X - M^k) \end{cases}, \quad (4)$$

where $M_1^{(0)}$ is an initial guess such as $D_\tau(X)$, $D_\tau(M^{(k-1)})$ is the operator of computing the singular value decomposition (SVD) of $M^{(k-1)}$ such that any singular value smaller than $\tau$ is replaced with 0. Unfortunately, both SDP solvers and the SVT algorithm are problematic when applied to a large-size data set [4]. Especially in the SVT, SVD computation is required at each iteration, which is time-consuming.

Another common approach to relax the rank is via matrix factorization, in which the unknown data matrix is expressed as the product of two low-rank matrices, $U$ and $V$ [4]. In this case, the low-rank condition is satisfied automatically. Function (1) can be transformed as:

$$\min_{U,V} \|\mathcal{P}_\Omega(X - UV^T)\|_F^2, \quad (5)$$

where $M = UV^T$.

The SRPCA method proposed in this paper can be grouped into this category. Compared with the first method, the matrix factorization-based approach performs much better on computation time. ALS is one of the popular matrix factorization-based methods, which originates from the power factorization method [33]. In the ALS algorithm, the observed entries are randomly partitioned into a number of subsets at first. Then, $U$ and $V$ are initialized through the SVD of the first subset of the observed matrix. Next, at each iteration when moving to the next subset, $U$ and $V$ are alternatively updated to minimize the difference between $UV^T$ and the observed entries of that subset. ALS decreases the computational time because it does not apply SVD at each iteration. However, it may lead to high inaccuracies at high levels of missing data, and it ignores the smoothness of the data set due to the random partitioning of the original matrix.

### B. Principal Component Analysis (PCA)

PCA is one of the most widely used statistical tools for data analysis and dimensionality reduction [34]. It has been applied in many different areas, such as quantitative finance [35], neuroscience [36], image processing [37], and so forth. PCA provides a roadmap for transforming the original data set to a new basis with a lower dimension, thus filtering out the noise and revealing the hidden simplified dynamics. Therefore, with PCA, it is possible to extract critically important information from original data, thus simplifying the data structure.

Suppose we have a data matrix $M \in \mathbb{R}^{m \times n}$. The goal of PCA is to find an orthonormal matrix where $P = MV$, such that the covariance matrix of $P$ is diagonalized. The covariance matrix of $P$ is expressed as:

$$S_P = \frac{1}{n-1} P^T P. \quad (6)$$

Because $P = MV$,

$$S_P = \frac{1}{n-1} (MV)^T (MV) = \frac{1}{n-1} V^T (M^T M) V. \quad (7)$$

Let $V$ be the eigenvectors matrix of $M^T M$; then matrix $S_P$ is diagonalized. This is because $M^T M = VDV^T$ and

$$\begin{aligned} S_P &= \frac{1}{n-1} V^T (VDV^T) V = \frac{1}{n-1} (V^T V) D (V^T V) \\ &= \frac{1}{n-1} (V^{-1} V) D (V^{-1} V) = \frac{1}{n-1} D. \end{aligned} \quad (8)$$

PCA is statistically intuitive and helps to reduce the data dimension; however, applying PCA iteratively for matrix completion is time-consuming. Therefore, in this paper, we initialize the matrices $P$ and $V$ via PCA and efficiently update them via the proposed algorithm in Section III.

## III. THE SPARSE REVERSE OF PCA (SRPCA)

### A. Problem Formulation

Based on the general matrix completion problem, the goal is to construct a matrix $M \in \mathbb{R}^{m \times n}$ that estimates the missing part of matrix $X \in \mathbb{R}^{m \times n}$. Let $\Omega = \{(i,j): x_{i,j} \text{ is observed}\}$, $\mathcal{P}_\Omega(X) \in \mathbb{R}^{m \times n}$ to be the matrix that preserves the entities in $\Omega$ and replaces the remaining entities by 0, $\Omega^\perp$ to be the complement of $\Omega$, and $\|.\|_F$ is the Frobenius norm. Following the matrix factorization approach, we focus on optimization problem (9) to find the matrix $M = PV^T$:

$$\min_{P,V} \|\mathcal{P}_\Omega(X - PV^T)\|_F^2. \quad (9)$$

where $P$ is the principal components matrix, and $V$ is the eigenvectors matrix of $M^T M$.

### B. Principal Components Estimation

To obtain the principal components, we first decompose the matrix $M^T M$:

$$M^T M = VDV^T, \quad (10)$$

where $D$ is a diagonal matrix with $\omega_j^2$ in its $j^{th}$ diagonal element, $\omega_j^2$ is the $j^{th}$ eigenvalue for $M^T M$, and $v_j$ is the $j^{th}$ eigenvector of $M^T M$ corresponding to the $j^{th}$ eigenvalue of $M^T M$.

Then, the principal components are estimated as:

$$P = MV_{1:r} = MU^T. \quad (11)$$

where $U^T = V_{1:r} \in \mathbb{R}^{r \times n}$.

Note that for computational efficiency, compression, and smoothing purposes, we consider the top $r$ eigenvectors. For matrix completion, $P$ and $U$ are updated iteratively.

### C. The SRPCA Algorithm

The first step in the proposed approach is data standardization, which is common in data analytics.

$$x_j = \frac{x_j - \mu_{X_{\Omega_j}}}{\sigma_{X_{\Omega_j}} + \epsilon} \forall j, \quad (12)$$

where $x_j$ is the $j^{th}$ column (i.e., $j^{th}$ sensor) of the matrix $X$, $\mu_{X_{\Omega_j}}$ and $\sigma_{X_{\Omega_j}}$ are the mean and standard deviation of the available elements in the $j^{th}$ column of the matrix $X$, and $\epsilon$ is a small positive value to avoid numerical instabilities when $\sigma_{X_{\Omega_j}} \to 0$.

Accordingly, an intuitive first approximation $M^{(0)}$ is

$$m_{(i,j)}^{(0)} = \begin{cases} x_{(i,j)} & (i,j) \in \Omega \\ N(0,1) & (i,j) \in \Omega^\perp \end{cases}. \quad (13)$$

Unlike many existing approaches, in this paper, each iteration starts with $M_\Omega^{(k)} = X_\Omega$ because the observed values of $X$ are unbiased estimates of the values in $\Omega$. This also serves as a reference point from which all iterations start. Then, we proceed from (13) to obtain the new updates for $P$

$$P^{(k)} = M^{(k)} U^{(k)T}. \quad (14)$$

and $M$

$$M^{(k+1)} = M^{(k)} U^{(k)T} U^{(k+1)} = P^{(k)} U^{(k+1)} \quad (15)$$

where $U^{(k+1)}$ is iteratively returned by the algorithm, which will be introduced next.

Those updates conclude some major advantages of the proposed SRPCA so far:

*(i) It starts with an unbiased estimate of the observed part of the matrix at every iteration.* This is critical for scenarios with high percentages of missing data, because the first few iterative updates of the matrix are highly dependent on unreliable random prior estimates of the missing part of the matrix. This may not only slow the convergence of the SRPCA but it may also lead to diverged estimates of the matrix $M$. Therefore, by keeping an unbiased estimate of the observed part of the matrix, it boosts the accuracy of the SRPCA to a certain extent.

*(ii) The new update $M^{(k+1)}$ is a smoother than the prior update $M^{(k)}$.* Therefore, the SRPCA also helps to smooth the original observed part of the matrix.

*(iii) The principal components are updated iteratively. This adds a layer of nonlinearity to the SRPCA.*

As shown in (15), the update $M^{(k+1)}$ depends on the updated eigenvectors $U^{(k+1)}$. Furthermore, the SRPCA updates $U^{(k+1)}$ to maintain a certain level of accuracy for the observed data.

$$U^{(k+1)} = \underset{U^{(k+1)}}{\operatorname{argmin}} \|\mathcal{P}_\Omega(X - P^{(k)} U^{(k+1)})\|_F^2. \quad (16)$$

The objective function in (16) aims to find $\mathcal{P}_\Omega(X - P^{(k)} U^{(k+1)}) \to 0$. This serves two purposes: (i) it ensures a smooth transition from $M_\Omega^{(k+1)} = \left(P^{(k)} U^{(k+1)}\right)_\Omega$ at the end of the iteration $k$ to $M_\Omega^{(k+1)} = X_\Omega$ at the beginning of the next iteration $k+1$, and (ii) it quantifies the differences between $M^{(k+1)}$ and the true matrix via $\|\mathcal{P}_\Omega(X - P^{(k)} U^{(k+1)})\|_F^2$ and aims to minimize the differences. Therefore, the updated $M^{(k+1)}$ is expected to provide a more realistic estimate of the missing data because now it provides a better estimate of the observed data.

Furthermore, because each column of $X$ can be expressed independently as a combination of the principal components, minimizing $\|\mathcal{P}_\Omega(X - P^{(k)} U^{(k+1)})\|_F^2$ is equivalent to

$$\min_{u_j^{(k)}} (x_j - P^{(k)} u_j^{(k+1)})^T W^{(j)} (x_j - P^{(k)} u_j^{(k+1)}), \forall j, \quad (17)$$

where $x_j$ is the $j$th column of $X$, $W^{(j)} \in \mathbb{R}^{m \times m}$ is the weight matrix for the $j^{\text{th}}$ column of $X$ and it is a diagonal matrix such that $w_{i,i}^{(j)} = 1$ if $(i,j) \in \Omega$ and 0 otherwise.

The sparse weight matrix $W^{(j)}$ provides all the weight on the observed values. Therefore, the solution of (18) is solely based on the observed part of $X$, and it can be written as the following:

$$u_j^{(k+1)} = \left(P^{(k)T} W^{(j)} P^{(k)}\right)^{-1} P^{(k)T} W^{(j)} x_j. \quad (18)$$

Applying (18) is scalable for big data in the presence of parallel computation capabilities. Intuitively, we can simultaneously compute different vectors of $U^{(k+1)}$. Furthermore, the weight matrices are sparse and they do not require full matrix operations; this is a built-in feature in many statistical software languages such as R and MATLAB.

Finally, the algorithm converges when the improvement between two successive iterations is smaller than a predefined tolerance threshold. In other words, the algorithm terminates when $\left\|\mathcal{P}_\Omega(X - P^{(k-1)} U^{(k)})\right\|_F^2 - \left\|\mathcal{P}_\Omega(X - P^{(k)} U^{(k+1)})\right\|_F^2 \leq \epsilon_{tol}$, where $M^{(k+1)} = P^{(k)} U^{(k+1)}$ and $\epsilon_{tol}$ is the tolerance threshold. Clearly, increasing $\epsilon_{tol}$ speeds up the algorithm convergence, but it also leads to a higher mean squared deviation $\left\|\mathcal{P}_\Omega(X - M^{(k+1)})\right\|_F^2$. Therefore, the choice of $\epsilon_{tol}$ depends on the application and the trade-off between speed and accuracy.

### D. Summary and Theoretical Findings for the SRPCA

Algorithm 1. The SRPCA for Matrix Completion

| 1 | $m_{(i,j)}^{(0)} = \begin{cases} x_{(i,j)} & (i,j) \in \Omega \\ N(0,1) & (i,j) \in \Omega^\perp \end{cases}$ | Data standardization and preprocessing |
|---|---|---|
| 2 | FOR $q = 1:1:n$ | |
| 3 | $\quad W^{(q)} = 0$ | Construct the sparse weight matrices once |
| 4 | $\quad w_{ii}^{(q)} = 1 \,\forall (i,q) \in \Omega$ | |
| 5 | END FOR LOOP | |
| 6 | $V^{(0)} D^{(0)} V^{(0)T} = M^{(0)T} M^{(0)}$ | Eigenvector decomposition |
| 7 | $U^{(0)} = V_{1:r}^{(0)T}$, $P^{(0)} = M^{(0)} U^{(0)T}$, $M^{(1)} = M^{(0)}$ | |
| 8 | FOR $k = 1:1:$maxIter | |
| 9 | $\quad$ Matrix smoothing | Optional |
| 10 | $\quad M_\Omega^{(k)} = X_\Omega$ | Key contribution |
| 11 | $\quad P^{(k)} = M^{(k)} U^{(k)T}$ | |
| 12 | $\quad$ FOR $j = 1:1:n$ | |
| 13 | $\quad\quad u_j^{(k+1)} = \left(P^{(k)T} W^{(j)} P^{(k)}\right)^{-1} P^{(k)T} W^{(j)} x_j$ | |
| 14 | $\quad$ END INNER FOR LOOP | |
| 15 | $\quad M^{(k+1)} = P^{(k)} U^{(k+1)}$ | Key contribution |
| 16 | $\quad$ IF $\left\|\mathcal{P}_\Omega(X - P^{(k-1)} U^{(k)})\right\|_F^2 - \left\|\mathcal{P}_\Omega(X - M^{(k+1)})\right\|_F^2 \leq \epsilon_{tol}$ | |
| 17 | $\quad\quad$ STOP & END OUTER FOR LOOP | |
| 18 | $\quad$ END IF CONDITION | |
| 19 | END OUTER FOR LOOP | |

Unlike some approaches in the literature, Lemma 1 shows that the performance of the SRPCA improves iteratively until it converges. This is a key finding because if the algorithm terminates for external reasons (e.g., computational time constraints), the algorithm output will be the best-calculated estimate until the unexpected termination.

**Lemma 1.** *The SRPCA converges iteratively with* $\left\|\mathcal{P}_\Omega(X - M^{(k+1)})\right\|_F^2 \leq \left\|\mathcal{P}_\Omega(X - P^{(k-1)} U^{(k)})\right\|_F^2$ (check Appendix A for details).

Lemma 2 provides an upper bound on the number of iterations until convergence, which also sets an upper bound on the computational time until convergence.

**Lemma 2.** *The SRPCA converges at an iteration* $K < \left\lceil \frac{\left\|\mathcal{P}_\Omega(X - P^{(0)} U^{(1)})\right\|_F^2}{\epsilon_{tol}} \right\rceil + 1$ (check Appendix A for details).

### E. Extension: The Fast SRPCA Algorithm

For many applications, the convergence rate is critical. It is often acceptable to converge to solutions that are close enough to optimality.

Recall that each iteration of the SRPCA starts with $M_\Omega^{(k)} = X_\Omega$ as a reliable unbiased estimate for the observed entities; however, this tends to slow down the convergence when $\left(P^{(k-1)} U^{(k)}\right)_\Omega$ is close but not equal to $X_\Omega$. Therefore, we propose the fast SRPCA for such case-studies:

$$M_\Omega^{(k)} = (1 - \alpha^*)\left(P^{(k-1)} U^{(k)}\right)_\Omega + \alpha^* X_\Omega, \quad (19)$$

where the initial value for $\alpha^* = 1$. Discussions on $\alpha^*$ are provided next.

There are two main advantages for the choice of (19). First, $\alpha^*$ serves as a step-size because $M_\Omega^{(k)} = \left(P^{(k-1)} U^{(k)}\right)_\Omega + \alpha^* \left(X_\Omega - \left(P^{(k-1)} U^{(k)}\right)_\Omega\right)$; therefore, it is expected that $\alpha^* \to 0$ when $\left(P^{(k-1)} U^{(k)}\right)_\Omega$ is close enough to $X_\Omega$. Second, $\alpha^*$ serves as a smoothing parameter for noisy datasets where $X_\Omega$ is a noisy estimate for the observed entries $\Omega$. For such noisy datasets, it is important to set $\alpha^* \to 0$ after enough iterations to avoid converging to a noisy estimate $M_\Omega^{(k)}$ that is close to $X_\Omega$.

**Lemma 3.** *If* $\alpha^* = 0$ *at the beginning of iteration* $K$, *the fast SRPCA converges at iteration* $K$ *with* $M^{(K+1)} = M^{(K)} = P^{(K)} U^{(K+1)} = P^{(K-1)} U^{(K)}$ (check Appendix B for details).

From Lemma 3, it is intuitive to define $\alpha^*$ as a decreasing function with respect to the iteration number $k$. This speeds the SRPCA convergence when $\left\|\mathcal{P}_\Omega(X - M^{(k+1)})\right\|_F^2$ converges slowly to $\left\|\mathcal{P}_\Omega(X - P^{(k-1)} U^{(k)})\right\|_F^2$. However, a random choice of $\alpha^*$ may result in an unreliable estimate even for the observed part of the matrix with a large error $\left\|\mathcal{P}_\Omega(X - M^{(k)})\right\|_F^2$.

Thus, the choice of $\alpha^*$ depends on $\left\|\mathcal{P}_\Omega(X - M^{(k)})\right\|_F$. Here, we propose $\alpha^*$ to be the solution for

$$\min_\alpha \left\|\mathcal{P}_\Omega(X - M^{(k)})\right\|_F + \lambda |\alpha|, \quad (20)$$

$\lambda$ is a tuning parameter and for this specific choice of the objective function, it is the convergence threshold for $\|\mathcal{P}_\Omega(X - M^{(k)})\|_F$ as shown in Lemma 4.

**Lemma 4.** *The closed form solution for (20) can be written as*
$$\alpha^* = \begin{cases} 0 & \|\mathcal{P}_\Omega(X - M^{(k)})\|_F \leq \lambda \\ 1 & \text{otherwise} \end{cases} \quad (21)$$
(check Appendix C for details).

Lemma 4 shows that the fast SRPCA sets $M_\Omega^{(k)} = X_\Omega$ only when $P^{(k)}U^{(k+1)}$ does not accurately reconstruct the observed part of the matrix (i.e., when $\|\mathcal{P}_\Omega(X - P^{(k)}U^{(k+1)})\|_F^2 > \lambda$). Note that for $\alpha^* = 1$ the Fast SRPCA becomes equivalent to the SRPCA in Algorithm 1 and for $\alpha^* = 0$ the Fast SRPCA terminates at the same iteration using Lemma 3.

Algorithm 2. The Fast SRPCA for Matrix Completion

| | | |
|---|---|---|
| 1 | $m_{(i,j)}^{(0)} = \begin{cases} x_{(i,j)} & (i,j) \in \Omega \\ N(0,1) & (i,j) \in \Omega^\perp \end{cases}$ | Data standardization and preprocessing |
| 2 | FOR $q = 1:1:n$ | |
| 3 | $W^{(q)} = 0$ | Construct the sparse weight matrices once |
| 4 | $w_{ii}^{(q)} = 1\ \forall (i,q) \in \Omega$ | |
| 5 | END FOR LOOP | |
| 6 | $V^{(0)}D^{(0)}V^{(0)T} = M^{(0)T}M^{(0)}$ | Eigenvector decomposition |
| 7 | $U^{(0)} = V_{1:r}^{(0)T}$, $P^{(0)} = M^{(0)}U^{(0)T}$, $M^{(1)} = M^{(0)}$ | |
| 8 | FOR $k = 1:1:$ maxIter | |
| 9 | Matrix smoothing if $\alpha^* \neq 0$ | Optional |
| 10 | $M_\Omega^{(k)} = (1-\alpha^*)(P^{(k-1)}U^{(k)})_\Omega + \alpha^* X_\Omega$ | |
| 11 | $P^{(k)} = M^{(k)}U^{(k)T}$ | |
| 12 | FOR $j = 1:1:n$ | |
| 13 | $u_j^{(k+1)} = (P^{(k)T}W^{(j)}P^{(k)})^{-1} P^{(k)T}W^{(j)}x_j$ | |
| 14 | END INNER FOR LOOP | |
| 15 | $M^{(k+1)} = P^{(k)}U^{(k+1)}$ | |
| 16 | $\alpha^* = \begin{cases} 0 & \|P_\Omega(X - M^{(k)})\|_F \leq \lambda \\ 1 & \text{otherwise} \end{cases}$ | |
| 17 | IF $\|\mathcal{P}_\Omega(X - P^{(k-1)}U^{(k)})\|_F^2 - \|\mathcal{P}_\Omega(X - M^{(k+1)})\|_F^2 \leq \epsilon_{\text{tol}}$ | |
| 18 | STOP & END OUTER FOR LOOP | |
| 19 | END IF CONDITION | |
| 20 | END OUTER FOR LOOP | |

## IV. APPLICATIONS

We validated the efficacy of the SRPCA approach on (i) a natural image, (ii) multisensor data, and (iii) a rating dataset. All analyses were done in MATLAB 9.2 using a laptop with a dual-core i7-6600U and 16 GB RAM.

### A. Image Matrix Completion

Images are often stored in the form of matrices, in which the intensity for pixel $(i,j)$ is stored in the matrix entry $(i,j)$. Furthermore, some pixels are often noisy or hard to obtain and it is common to use matrix completion to reconstruct images. Here, the algorithm is validated on a natural image (475x344). Specifically, a randomly selected subset (50%, 70%, and 80%) of the pixels are removed and the SRPCA is then applied to reconstruct the image with $\epsilon_{\text{tol}} = 10^{-4}$.

The benchmark methods used for comparison are: (i) inexact augmented Lagrange multiplier (**ALM**) [23], (ii) singular value thresholding (**SVT**) [1], (iii) variational Bayesian low-rank (**VBLR**) [38], (iv) nuclear norm with templates first-order conic solvers (**TFOCS**) [39], and (v) **ALS** built-in MATLAB.

Table 1. The computational time and the error $\|(X - M^{(k+1)})\|_F^2$ for 100 replications at different levels of missing data

| | Time (mean/std) | | | $10^3$ x Error (mean/std) | | |
|---|---|---|---|---|---|---|
| | 50% | 70% | 80% | 50% | 70% | 80% |
| SRPCA | **0.92** 0.046 | **0.65** 0.038 | **0.73** 0.056 | **9.6** 0.25 | **29.2** 0.60 | **51.6** 1.88 |
| ALM | 3.54 0.099 | 2.08 0.15 | 1.83 0.13 | 11.7 0.26 | 52.2 1.44 | 128 4.07 |
| SVT | 2.69 0.089 | 2.66 0.10 | 2.68 0.19 | 22.1 0.51 | 94.9 2.32 | 216 6.22 |
| VBLR | 150 2.18 | 24.22 0.36 | 7.09 0.22 | 20.9 1.12 | 37.2 1.21 | 66.6 3.19 |
| TFOCS | 9.50 1.45 | 4.46 0.52 | 2.96 0.45 | 22.4 0.51 | 95.6 2.32 | 217 6.21 |
| ALS | 16.85 0.50 | 8.56 0.43 | 10.3 0.68 | 63.0 3.61 | 155 11.6 | 573 43.9 |

The computational time required to reconstruct the image and the mean squared difference $\|X - M\|_F^2$ is shown in Table 1. Also, the reconstructed images based on the proposed and benchmark methods are shown in Figure 1.

From Figure 1 and Table 1, we can see that (i) the error increases with the increase in missing data, (ii) the error from the SRPCA is smaller than that from the benchmark methods at all levels of missing data, and (iii) the computational time of the SRPCA is smaller than that of the benchmark at all levels of missing data. The first observation is natural because the matrix rank is probably underestimated with less observed data, leading to higher errors. The second and third observations show the efficacy of the proposed SRPCA over the benchmark methods. This is expected because the proposed approach considers the smoothness of the matrix and efficiently updates the principal components and eigenvectors without explicitly running the eigenvector decomposition.

### B. Multi-task Learning: Multisensor Matrix Completion

With the rapid development of sensor technology and wireless communication, it is now ideally possible to monitor different components and characteristics of complex systems in near real-time. However, in real-life applications, the ideal case is always met with technical difficulties such as wireless shutdowns, slow communication with data centers, delayed responses, limited sensor capabilities, poor sensor quality, and others. This occasionally raises issues similar to (i) ignoring some of the incoming data from some sensors, (ii) not receiving data from some sensors, and (iii) receiving noisy data. Note that in the field of multisensor data, data analysis is often conducted periodically; therefore, the SRPCA provides a handy tool to update missing data before the analysis or upon request.

In this section, we focus on a well-known published dataset on aircraft gas turbine engines [40]. The dataset contains 100 engines that ran until failure. For each of those engines,

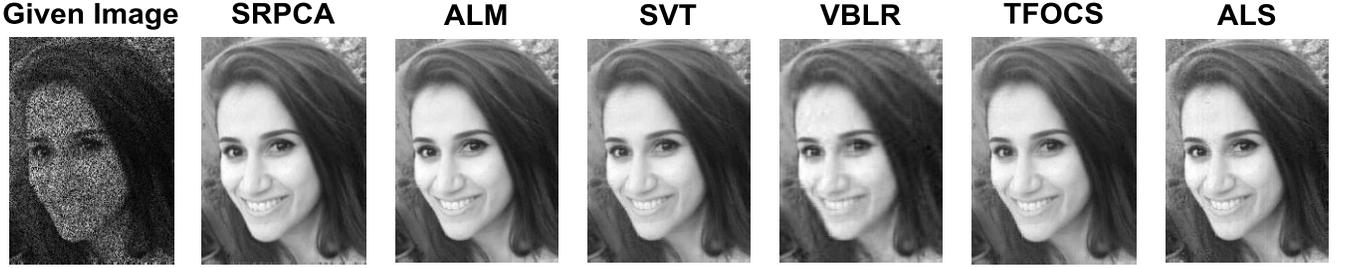
Figure 1a. Reconstructed images with 50% missing pixels (the rank for SRPCA, VBLR and ALS is set to 80)

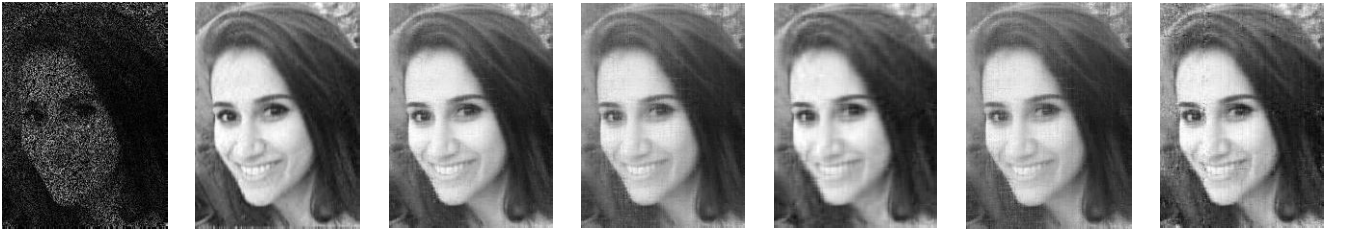
Figure 1b. Reconstructed images with 70% missing pixels (the rank for SRPCA, VBLR and ALS is set to 40)

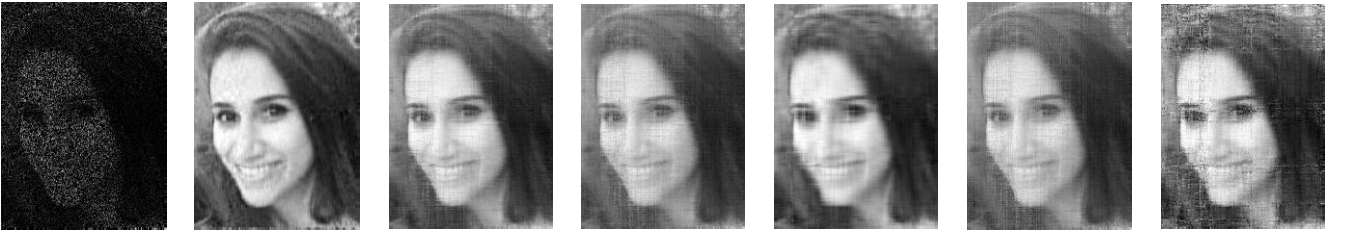
Figure 1c. Reconstructed images with 80% missing pixels (the rank for SRPCA, VBLR and ALS is set to 30)

measurements from 21 sensors were recorded after each cycle with no missing data. Overall, the dataset contains 20631 observations from each of the 21 sensors. For this specific dataset, 12 sensors are often selected in the literature due to their consistent trends across the 100 engines. For the purpose of this paper, we focus on those 12 sensors.

To validate the SRPCA, we first stack the multisensor data from all 100 engines in one giant matrix. Then, we randomly remove different percentages (10%, 30%, 50%, and 70%) of the giant matrix. Note that we considered all 100 engines to validate that the SRPCA is scalable. We also set the rank for the SRPCA, ALS, and VBLR to be 1.

Figure 2 shows the results from a specific scenario where 70 percent of the data is removed. Additionally, Table 2 summarizes the computational time and the mean squared-difference between the SRPCA predictions and the true original matrix ($\|X - M\|_F^2$).

From Figure 2, we can see that even at a high level of missing data (70%), the SRPCA predictions for every sensor are accurate. This shows the efficacy of the SRPCA in multisensor data environments. Furthermore, the reconstructed signals are smoother than the original signals, which shows the smoothing capabilities of the SRPCA.

Table 2 shows that the deviations between the original matrix and the reconstructed matrix are acceptable. Note that the reconstructed matrix is a smoother for the original matrix as shown in Figure 2, and therefore differences between both matrices are expected. The table also shows that the computational time is almost similar at different levels of missing data. Finally, we can see that the computational time under all tested levels of missing data is considerably low; this provides another valuable feature for the SRPCA and its applicability for near real-time analyses.

Table 2. The SRPCA performance under different levels of missing data (100 replications/missing level) with $\epsilon_{\text{tol}} = 10^{-4}$

|     | Time (secs) $(\mu_t, \sigma_t)$ | | $\|X - M\|_F^2$ $(\mu_d, \sigma_d)$ | |
| --- | --- | --- | --- | --- |
| 10% | 0.098 | 0.0095 | 0.262 | 0.0003 |
| 30% | 0.096 | 0.0041 | 0.270 | 0.0006 |
| 50% | 0.088 | 0.0061 | 0.279 | 0.0008 |
| 70% | 0.096 | 0.0031 | 0.295 | 0.0014 |

Table 3. The mean computational times for the benchmark methods for 100 replications

|     | Time (secs) (mean/std) | | | | |
| --- | --- | --- | --- | --- | --- |
|     | ALM | SVT | VBLR | TFOCS | ALS |
| 10% | 3.000 | 0.560 | 3.437 | 1.200 | 17.20 |
|     | 0.131 | 0.034 | 0.133 | 0.080 | 2.313 |
| 30% | 2.560 | 0.656 | 3.324 | 1.127 | 19.55 |
|     | 0.170 | 0.065 | 0.290 | 0.102 | 3.247 |
| 50% | 2.002 | 0.665 | 2.797 | 0.973 | 24.78 |
|     | 0.267 | 0.103 | 0.393 | 0.142 | 8.189 |
| 70% | 1.665 | 0.745 | 1.910 | 0.993 | 35.16 |
|     | 0.205 | 0.100 | 0.218 | 0.123 | 15.53 |

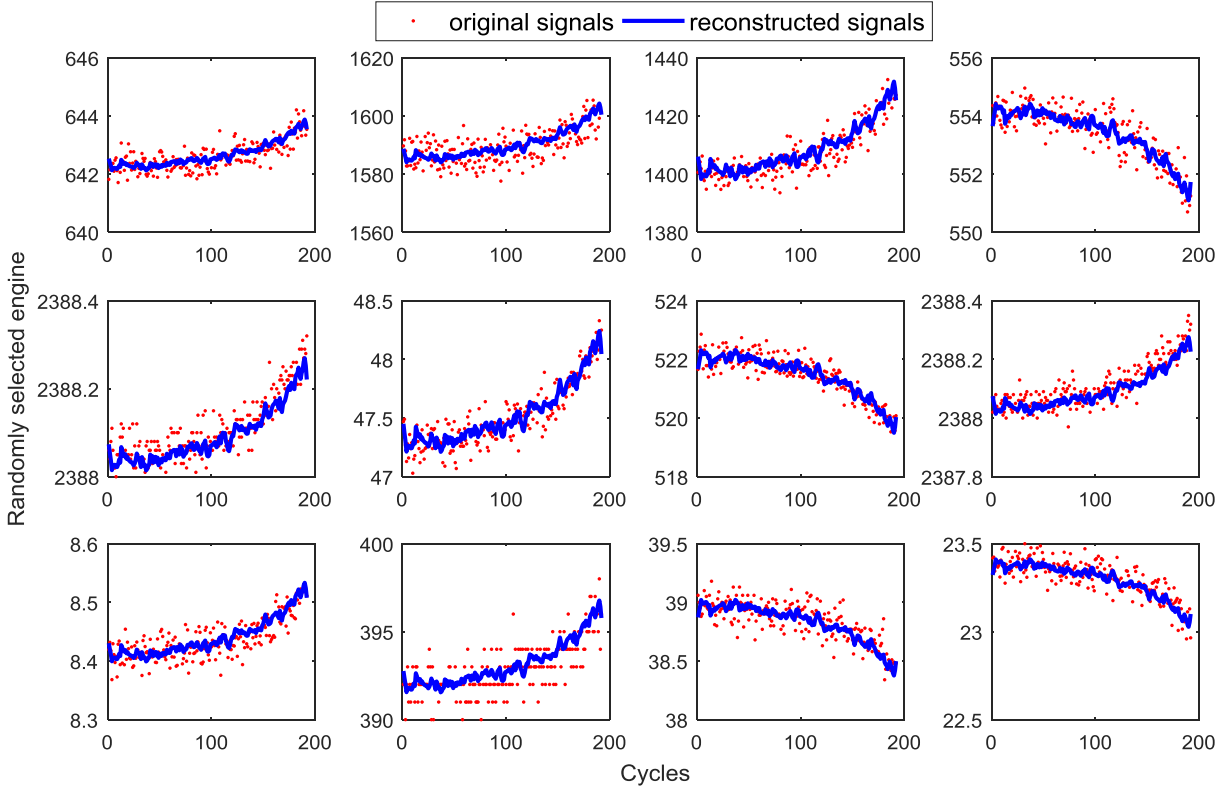

Figure 2. Reconstructed multisensor signals via the SRPCA approach with 70% missing data

The computational times for the benchmark methods are summarized in Table 3. We did not consider error-based metrics for comparison because the dataset is noisy and differences from the original data are expected. From the table, we can see that the SRPCA is much faster than the benchmark methods and that it is hard to leverage the benchmark methods for near real-time analyses.

As a conclusion, this example shows the efficacy, the efficiency, and the scalability of the SRPCA for multisensor data. For future studies, the SRPCA can be coupled with reliability analysis, degradation modeling, and predictive maintenance to increase the value of complex systems.

*C. Recommender Systems: MovieLens 100k*

Finally, we evaluate the performance of the SRPCA for the movie recommendations in the MovieLens 100k dataset that is available at https://grouplens.org/datasets/movielens/. The dataset contains 100k recommendations from 943 users for 1682 movies, which can be represented in a matrix of size 1682x943. We analyze scenarios where 50% and 80% of the recommendations are randomly removed. The performance is evaluated by the normalized mean absolute error metric (NMAE) in [41]. The results are summarized in Table 4.

$$\text{NMAE} = \frac{\sum_{(i,j)\in\mathbf{\Omega}^\perp}|m_{i,j} - x_{i,j}|}{(x_{\max} - x_{\min})|\mathbf{\Omega}^\perp|}, \quad (22)$$

where $x_{\max}$ and $x_{\min}$ are the maximum and minimum values, respectively, for the available recommendations.

Table 4 shows that the VBLR is the most accurate and the SRPCA comes next; however, the SRPCA is the fastest and the VBLR comes next. Also, from the two applications mentioned previously, the VBLR seems to be appropriate for extremely low-rank matrices, and its computational time increases exponentially with the rank. In conclusion, this example shows that the proposed SRPCA performs with similar accuracy to the benchmark methods but at a faster rate.

Table 4. The mean computational time and mean NMAE for 100 replications with $\epsilon_{\text{tol}} = 10^{-3}$

|     | Time (secs)/NMAE | | | | | |
| --- | --- | --- | --- | --- | --- | --- |
|     | SRPCA | ALM | SVT | VBLR | TFOCS | ALS |
| 50% | **2.077** | 14.04 | 9.601 | 3.894 | 18.15 | 4.946 |
|     | 0.179 | 0.186 | 0.217 | **0.177** | 0.208 | 0.549 |
| 80% | **2.007** | 14.17 | 9.167 | 3.729 | 16.18 | 4.785 |
|     | 0.180 | 0.186 | 0.221 | **0.177** | 0.214 | 0.553 |

V. CONCLUSIONS AND FUTURE WORK

The rapid development of sensing technologies has provided unprecedented environments of big data that can be leveraged for reliability analysis, collaborative filtering, predictive analytics, and transfer learning. However, such data environments are often sparse due to data storage constraints, adaptive sampling rates, connectivity issues, poor calibrations, high sensitivities, etc. Therefore, there is always a constant demand for matrix completion approaches in many domains to accurately estimate missing or unavailable data. In this paper, we propose a novel matrix completion approach for incomplete datasets; the contribution of this paper is multifold:

*(i) This approach maintains a certain level of smoothness across the matrix.*

*(ii) It converges iteratively, which is critical for scenarios that terminate the algorithm before convergence.*

*(iii) It is computationally tractable, with a controlled upper bound on the number of iterations until convergence, which is critical in the presence of computational constraints.*

The above-mentioned contributions provide confidence that the predicted values of the missing data are reliable estimates of the true values of the missing data. Additionally, the efficacy of the SRPCA algorithm is validated on a natural image, multisensor dataset, and rating dataset. For future studies, it is important to integrate matrix partitioning techniques with the SRPCA to achieve better computational efficiency. It is also important to develop stochastic extensions to the SRPCA.

## APPENDIX A

This appendix proves Lemmas 1 and 2. First, the rationale of adding $M_\Omega^{(k)} = X_\Omega$ is expected to improve the performance iteratively for scenarios where $\|\mathcal{P}_\Omega(X - P^{(k-1)}U^{(k)})\|_F^2$ is large. Intuitively, the algorithm should terminate when $\|\mathcal{P}_\Omega(X - P^{(k-1)}U^{(k)})\|_F^2 - \|\mathcal{P}_\Omega(X - P^{(k)}U^{(k+1)})\|_F^2 \leq \epsilon_{tol}$, meaning that the error term $\|\mathcal{P}_\Omega(X - P^{(k-1)}U^{(k)})\|_F^2$ is not significantly decreasing anymore.

Next, we show that $\|\mathcal{P}_\Omega(X - P^{(k-1)}U^{(k)})\|_F^2 - \|\mathcal{P}_\Omega(X - P^{(k)}U^{(k+1)})\|_F^2$ will keep decreasing until it becomes smaller than $\epsilon_{tol}$.

Specifically, at each iteration $k$, it is either
$$\|\mathcal{P}_\Omega(X - P^{(k-1)}U^{(k)})\|_F^2 \leq \|\mathcal{P}_\Omega(X - P^{(k)}U^{(k+1)})\|_F^2 + \epsilon_{tol}$$
or
$$\|\mathcal{P}_\Omega(X - P^{(k)}U^{(k+1)})\|_F^2 < \|\mathcal{P}_\Omega(X - P^{(k-1)}U^{(k)})\|_F^2 - \epsilon_{tol}.$$

For $\|\mathcal{P}_\Omega(X - P^{(k-1)}U^{(k)})\|_F^2 \leq \|\mathcal{P}_\Omega(X - P^{(k)}U^{(k+1)})\|_F^2 + \epsilon_{tol}$, then the algorithm terminates at iteration $k$ by satisfying $\|\mathcal{P}_\Omega(X - P^{(k-1)}U^{(k)})\|_F^2 - \|\mathcal{P}_\Omega(X - P^{(k)}U^{(k+1)})\|_F^2 \leq \epsilon_{tol}$.

For $\|\mathcal{P}_\Omega(X - P^{(k)}U^{(k+1)})\|_F^2 < \|\mathcal{P}_\Omega(X - P^{(k-1)}U^{(k)})\|_F^2 - \epsilon_{tol}$, then $\|\mathcal{P}_\Omega(X - P^{(k)}U^{(k+1)})\|_F^2 < \|\mathcal{P}_\Omega(X - P^{(k-1)}U^{(k)})\|_F^2$ because $\epsilon_{tol} > 0$, and we move to the next iteration $k + 1$.

Similarly, either the SRPCA terminates at $k + 1$ or $\|\mathcal{P}_\Omega(X - P^{(k+1)}U^{(k+2)})\|_F^2 < \|\mathcal{P}_\Omega(X - P^{(k)}U^{(k+1)})\|_F^2 - \epsilon_{tol} < \|\mathcal{P}_\Omega(X - P^{(k)}U^{(k+1)})\|_F^2 < \|\mathcal{P}_\Omega(X - P^{(k-1)}U^{(k)})\|_F^2$.

This concludes the fact that $\|\mathcal{P}_\Omega(X - P^{(k+1)}U^{(k+2)})\|_F^2$ decreases iteratively until $\|\mathcal{P}_\Omega(X - P^{(k-1)}U^{(k)})\|_F^2 - \|\mathcal{P}_\Omega(X - P^{(k)}U^{(k+1)})\|_F^2 \leq \epsilon_{tol}$ is satisfied. This concludes the proof for Lemma 1.

Finally, we show that there exists an iteration $K$ such that $\|\mathcal{P}_\Omega(X - P^{(K-1)}U^{(K)})\|_F^2 - \|\mathcal{P}_\Omega(X - P^{(K)}U^{(K+1)})\|_F^2 \leq \epsilon_{tol}$. Assume that the algorithm did not converge at iteration $K - 1$; therefore, $0 \leq \|\mathcal{P}_\Omega(X - P^{(K-1)}U^{(K)})\|_F^2 < \|\mathcal{P}_\Omega(X - P^{(K-2)}U^{(K-1)})\|_F^2 - \epsilon_{tol} < \|\mathcal{P}_\Omega(X - P^{(K-3)}U^{(K-2)})\|_F^2 - 2\epsilon_{tol} < \cdots < \|\mathcal{P}_\Omega(X - P^{(0)}U^{(1)})\|_F^2 - (K - 1)\epsilon_{tol}$. However, for those inequalities to hold, we must have $\|\mathcal{P}_\Omega(X - P^{(1)}U^{(2)})\|_F^2 - (K - 2)\epsilon_{tol} > 0$. Thus, the algorithm will terminate at an iteration $K < \left\lceil \frac{\|\mathcal{P}_\Omega(X - P^{(0)}U^{(1)})\|_F^2}{\epsilon_{tol}} \right\rceil + 1$ satisfying $\|\mathcal{P}_\Omega(X - P^{(K-1)}U^{(K)})\|_F^2 - \|\mathcal{P}_\Omega(X - P^{(K2)}U^{(K+1)})\|_F^2 \leq \epsilon_{tol}$. This concludes the proof for Lemma 2.

## APPENDIX B

This appendix proves Lemma 3. If $\alpha^* = 0$ at the beginning of iteration $K$, then the solution for
$$\min_{u_j^{(K+1)}} (x_j - P^{(K)}u_j^{(K+1)})^T W^{(j)} (x_j - P^{(K)}u_j^{(K+1)})$$
$$\forall j = 1, \ldots n,$$
is
$$u_j^{(K+1)} = \left(P^{(K)T}W^{(j)}P^{(K)}\right)^{-1} P^{(K)T}W^{(j)}x_j \quad \forall j = 1, \ldots n.$$

This is also the solution for
$$\min_{u_j^{(K+1)}} \left(x_j - M^{(K)}U^{(K)T}u_j^{(K+1)}\right)^T W^{(j)} \left(x_j - M^{(K)}U^{(K)T}u_j^{(K+1)}\right)$$
$$\forall j = 1, \ldots n,$$
which can be written as
$$\min_{u_j^{(K+1)}} \left(x_j - P^{(K-1)}U^{(K)}U^{(K)T}u_j^{(K+1)}\right)^T W^{(j)} \left(x_j - P^{(K-1)}U^{(K)}U^{(K)T}u_j^{(K+1)}\right)$$
$$\forall j = 1, \ldots n,$$
because $\alpha^* = 0$.

Therefore,
$$U^{(K)}U^{(K)T}u_j^{(K+1)} = \left(P^{(K-1)T}W^{(j)}P^{(K-1)}\right)^{-1} P^{(K-1)T}W^{(j)}x_j$$
$$= u_j^{(K)}$$
$$\forall j = 1, \ldots n,$$

Finally, multiplying both sides of the equation by $P^{(K-1)}$:
$$P^{(K-1)}U^{(K)}U^{(K)T}u_j^{(K+1)} = P^{(K-1)}u_j^{(K)} \quad \forall j = 1, \ldots n.$$
$$M^{(K)}U^{(K)T}u_j^{(K+1)} = m_j^{(K)} \quad \forall j = 1, \ldots n \text{ because } \alpha^* = 0.$$
$$P^{(K)}u_j^{(K+1)} = m_j^{(K+1)} \quad \forall j = 1, \ldots n.$$
$$m_j^{(K+1)} = m_j^{(K)} = P^{(K-1)}u_j^{(K)} \quad \forall j = 1, \ldots n \text{ because } \alpha^* = 0.$$

Thus, the fast SRPCA, in Algorithm 2, converges at the end of iteration $K$ by satisfying the $\left\|\mathcal{P}_{\Omega}(X - P^{(K-1)}U^{(K)})\right\|_F^2 - \left\|\mathcal{P}_{\Omega}(X - M^{(K+1)})\right\|_F^2 = 0 < \epsilon_{\text{tol}}$.

APPENDIX C

This appendix proves Lemma 4. First, we rewrite
$$\min_{\alpha} \left\|\mathcal{P}_{\Omega}(X - M^{(k)})\right\|_F + \lambda |\alpha|$$
as
$$\min_{\alpha} \left\|\mathcal{P}_{\Omega}(X - (1-\alpha)P^{(k-1)}U^{(k)} - \alpha X)\right\|_F + \lambda |\alpha|$$
$$= \min_{\alpha} \left\|(1-\alpha)\mathcal{P}_{\Omega}(X - P^{(k-1)}U^{(k)})\right\|_F + \lambda |\alpha|$$
$$= \min_{\alpha} |1-\alpha| \left\|\mathcal{P}_{\Omega}(X - P^{(k-1)}U^{(k)})\right\|_F + \lambda |\alpha|.$$

Knowing that $\left\|\mathcal{P}_{\Omega}(X - P^{(k-1)}U^{(k)})\right\|_F > 0$ and $\lambda \geq 0$, then the solution satisfies $0 \leq \alpha^* \leq 1$. Specifically, if
$$\left\|\mathcal{P}_{\Omega}(X - P^{(k-1)}U^{(k)})\right\|_F < \lambda$$
then $\left\|\mathcal{P}_{\Omega}(X - P^{(k-1)}U^{(k)})\right\|_F$
$$= \min_{\alpha} |1-\alpha| \left\|\mathcal{P}_{\Omega}(X - P^{(k-1)}U^{(k)})\right\|_F + \lambda |\alpha|$$
with $\alpha^* = 0$; otherwise if
$$\left\|\mathcal{P}_{\Omega}(X - P^{(k-1)}U^{(k)})\right\|_F > \lambda$$
then $\lambda = \min_{\alpha} |1-\alpha| \left\|\mathcal{P}_{\Omega}(X - P^{(k-1)}U^{(k)})\right\|_F + \lambda |\alpha|$
with $\alpha^* = 1$.

For the case $\left\|\mathcal{P}_{\Omega}(X - P^{(k-1)}U^{(k)})\right\|_F = \lambda$, any value $0 \leq \alpha^* \leq 1$ is a valid solution, but we chose $\alpha^* = 0$ to speed up the convergence of the algorithm.

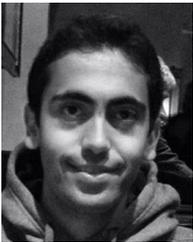

**Abdallah Chehade (M–2018)** received the B.S. degree in mechanical engineering from the American University of Beirut, Beirut, Lebanon, in 2011 and the M.S. degree in mechanical engineering, the M.S. degree in industrial engineering, and the Ph.D. in industrial engineering from the University of Wisconsin-Madison in 2014, 2014, and 2017, respectively. Currently, he is an assistant professor in the Department of Industrial and Manufacturing Systems Engineering at the University of Michigan-Dearborn. His research interests are data-driven models, data fusion for process modeling, and optimization of data-analysis. Dr. Chehade is a member of INFORMS, IEEE and IISE.

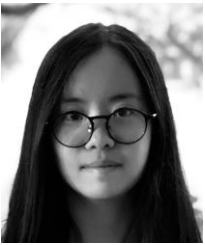

**Zunya Shi** received the B.S. degree in business administration and the M.S. degree in industrial engineering from Xi'an Jiaotong University, Xi'an, China, in 2014 and 2016, respectively. At present, she is working toward the Ph.D. degree in the Department of Industrial and Manufacturing Systems Engineering at the University of Michigan-Dearborn. Her research interests focus on prognosis and degradation analysis. She is a member of IISE.